\documentclass[11pt,a4paper]{article}
\usepackage{naacl2021}
\usepackage{times}
\usepackage{tabularx}
\usepackage{graphicx}
\usepackage{latexsym}
\usepackage{amsmath}

\usepackage{microtype}

\title{An Attention Based Neural Network for Code Switching Detection: English \& Roman Urdu}

\author{Aizaz Hussain, Muhammad Umair Arshad\\
  Artificial Intelligence \& Machine Learning lab\\
  National University of Computer and Emerging Sciences, FAST\\
  Islamabad, Pakistan \\
  \texttt{\{aizaz.hussain,umair.arshad\}@nu.edu.pk} \\
  }

\date{}

\begin{document}
\maketitle
\begin{abstract}
 

Code-switching is a common phenomenon among people with diverse lingual background and is widely used on the internet for communication purposes. In this paper, we present a  Recurrent Neural Network combined with the Attention Model for Language Identification in Code-Switched Data in English and low resource Roman Urdu. The attention model enables the architecture to learn the important features of the languages hence classifying the code switched data. We demonstrated our approach by comparing the results with state of the art models i.e. Hidden Markov Models, Conditional Random Field and Bidirectional LSTM. The models evaluation, using confusion matrix metrics, showed that the attention mechanism provides improved the precision and accuracy as compared to the other models.  

\end{abstract}

\section{Introduction}

In sociolinguistics, a language is referred to as code. Using two languages to convey meaning is referred to as Code-Mixing (CM) and Code-Switching (CS). In this age of internet and information technology, information exchange has grown exponentially. People around the globe are connected through the internet and all these people have diverse lingual backgrounds. Using the support of two languages to convey meaning is a common phenomenon nowadays. 

Code-Mixing is the most used method to communicate with one another on social media platforms. Code-Mixing means that in a single sentence a person uses multilingual vocabulary, grammar for the delivery of meaningful information. For example, consider a Pakistani student whose native language is Urdu but in school, the communication mode is in English. For such students to communicate it is commonly observed that they mix English with Urdu for the part they find difficult to speak in English. Code-switching is used when you have to switch to other languages to deliver messages. For example, a teacher in a class teaching a foreign language will first speak in the target language and then will make use of the native language to help students understand the meaning of the sentence.

Code switching poses a significant problem for text mining algorithms and online social platforms. These algorithms fail when it comes to multilingualism text processing. Existing tools are limited to the extraction of semantics for a single or one-to-one language transliteration. Our paper makes the following contributions: 1) We show that extraction with the use of attention model improves the accuracy 2) Code mixing language identificaiton for low resource language i.e. English and Roman Urdu 3) We present a new public dataset to support research on code mixed language identification on Roman Urdu and English

In this paper, we propose to use the transformer-based neural network for the detection of Multilingual text in English and Urdu from the given dataset. We will compare the results of the proposed architecture with different classification models i.e. Hidden Markov Models (HMM), Conditional Random Fields (CRF), and a BiDirectional LSTM. The rest of the paper is divided into the following 4 sections. In section 1, we will discuss what related work has been done for the detection of multilingual language detection until now. In section 3, the proposed methodology will be explained in detail along with the working of pre-existing models. Section 4, explains the complete setup of the dataset, experimentations and comparison of the results. In section 5, we will finally summarize the findings of this paper.

\section{Related Work}
The problem of multilingual language detection from the text has been long identified for 30 years \cite{joshi-1982-processing, naeem2020subspace}. With the classification of multilingual text, much useful semantics can be extracted from the text \cite{qamarrelationship, awan2021top}. The first step in this domain is to identify the POS tags \cite{nguyen2018improved, majeed2020emotion, zahid2020roman} in multilingual text.
From the machine learning perspective, one can view multilingual text detection as a classification problem. Machine learning algorithms have been applied to classify two or more languages \cite{amini2009learning, naeem2020deep, beg2013constraint,uzair2019weec}. Language identification from code mixed data for English Telugu languages showed that applying Hidden Markov Model (HMM), Support Vector Machines (SVM), and Conditional Random Field (CRF), the CRF outperformed the rest of the algorithms \cite{gundapu2020word}. Similarly, if the context is taken into consideration with traditional machine learning models, as proposed in \cite{nguyen2013word}, an accuracy of 98\%  was achieved for the Turkish and dutch languages dataset. 

Each language has different pronunciations and phonology combinations. So with the use of deep neural networks the proposition is to learn the morphemes and phonemes to predict the probability of each word belonging to a specific language. 
In the paper \cite{jurgens2017incorporating}, the authors proposed to solve the multilingual language identification problem by using neural networks with the to learn the character combinations. In the '90s, LSTM \cite{hochreiter1997long} was introduced to help preserve the context for long and short memory, which is being used in latest state preserving AI in games \cite{zafar2019constructive, zafar2018deceptive, zafar2019using}. LSTM's have been applied to many text processing problems including multilingual text detection \cite{samih2016multilingual} thus achieving an accuracy of 0.83 and 0.90 on MSA - Egyptian and Spanish - English languages.

Urdu is the widely spoken language in the south Asian region. 
Detection of sentiments from Urdu text gives very useful insights \cite{mehmood2019sentiment}, especially in the context of local businesses. 
In the paper \cite{amjad2020data}, the author using machine translation between English-Urdu detects the fake news. A lot of work has been done in terms of basic language processing for Urdu i.e. Creation of benchmark resource for processing of Urdu language \cite{hussain2008resources, arshad2019corpus}, POS tagging \cite{muaz2009analysis}, Stemming \cite{khan2012light} etc.

\begin{figure*}
\centering
\centering
     \includegraphics[width=0.99\linewidth]{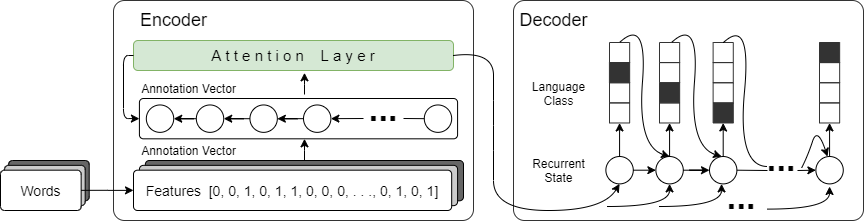}
     \caption{\label{Fig:image05}The architecture contains of two parts i.e encoder and decoder. The encoder converts the words into features and then the attention layer assigns the weights to the respective features. In the decoder part, the values from the attention layer are passed into the recurrent network thus learning the sequence of the word \& classifying~it.}
\end{figure*}



Roman Urdu is a complex form of Urdu language communication. There are many inconsistencies which are hard to decipher when it comes to machine however it is easy for humans to understand the meaning of a particular word written in Roman Urdu. So the very first task needed was the correct depiction of Roman Urdu words in Urdu. 
Another problem with Roman Urdu is that it is also written in Roman characters which are used for English also, which makes it difficult to differentiate between multilingual text.

\section{Methodology}
    
\subsection{Recurrent Neural Network:}
\vspace{3pt}

\noindent Recurrent Neural Networks are good for processing sequential data. Text is also sequential data with a combination of many characters. RNN's are reliable for short sequences. Considering a simple feed-forward neural network that has a single input layer, a hidden layer, and an output layer. The information is saved by adding the loop in the hidden state. This allows us to make a flow of information between different steps. The hidden state is updated with every iteration given the old hidden state as in input to the new one. The hidden state is updated on the bases of the following relation:
\begin{align}\label{eq1}
    \textbf{\textit{h\textsubscript{t}}} = f\textsubscript{W}(x\textsubscript{t}, h\textsubscript{t - 1})
\end{align}

Whereas, \textit{h\textsubscript{t-1}} in equation \ref{eq1} represents the old state of \textit{h\textsubscript{t}} which is passed to update the knowledge of the network.

\vspace{5pt}
\subsection{Attention Model:}\label{secattentionmodel}
\vspace{3pt}

\noindent The attention mechanism, Rather than building a single context vector the idea is to create shortcuts between the context vector and the entire source input. In this way only important features from the context vectors are highlighted. Suppose we have sequence \textbf{\emph{x}} and target sequence \textbf{\textit{y}}, such that:
\begin{align}\label{eq2}
     \textit{x} & = [x\textsubscript{1}, x\textsubscript{2}, x\textsubscript{3}, . . . , x\textsubscript{n}] \nonumber\\
     \textit{y} & = [y\textsubscript{1}, y\textsubscript{2}, y\textsubscript{3}, . . . ,y\textsubscript{n}] \nonumber \\
     \text{whereas, } \textbf{\textit{h\textsubscript{\textit{i}}}} & = [\textit{h\textsubscript{\textit{i}}}]\textsuperscript{T}, \textbf{\textit{  i}} = 1,2, ..., \textit{n}
\end{align}

\textbf{\textit{h\textsubscript{i}}} represents the hidden state in the encoder. The decoder contains the hidden state \textbf{\textit{s\textsubscript{t}}}, as shown in eq \ref{eq3} where t is the position of the output word. 

\begin{align}\label{eq3}
    \textbf{\textit{s\textsubscript{t}}} & = f({\textit{s\textsubscript{t-1}}}, \textit{y\textsubscript{t-1}}, \textit{c\textsubscript{t}}) \\
    \text{where,  } \textbf{\textit{c\textsubscript{t}}} & = \sum_{i=1}^{n} \alpha\textsubscript{\textit{t, i}}\textit{h\textsubscript{i}} \nonumber
\end{align}

The set of {$\alpha\textsubscript{t, i}$} are weights for each output for the emphasis on each source hidden state.

\vspace{5pt}
\subsection{Recurrent Neural Network with Attention Model:}
\vspace{3pt}

\noindent\textbf{Feature Extraction:}\label{featureext}
\vspace{3pt}

\noindent To extract features from words we used skipgram model using Fasttext library. At first the word is passed converted in a vector of 300 dimensions using one hot encoding. A dot product between the word \textbf{\textit{w}} features and weights on $|v|$ neurons is performed. This hidden layer do not have any activation function. The output vector H[1, N] is then taken dot product with W[N, V] and will give vector \textbf{$\overrightarrow{U}$} as output. Using softmax function the probability for each vector is calculated. The vector with the highest probability represents as the output.

\vspace{5pt}
\noindent\textbf{Model:}
\vspace{3pt}

\noindent We propose to combine a recurrent neural network combined with the attention model to efficiently enable the network to learn the features of each language. As explained in section \ref{secattentionmodel}, the network contains two part i.e. an encode and a decoder. The features extracted from words \ref{featureext} are passed into the input layer. This input layer is passed into a bidirectional Recurrent Neural Network. In a bi-directional RNN the sequence is first passed in the order {\textbf{\textit{S}} = $\{s\textsubscript{1}, s\textsubscript{2}, s\textsubscript{3}, ..., s\textsubscript{n}\}$} and then in the other layer it is passed the in opposite direction {\textbf{\textit{S\textsuperscript{'}}}~=~ $\{s\textsubscript{n}, s\textsubscript{n-1}, s\textsubscript{n-2}, ..., s\textsubscript{1}\}$}.The results from both the layer are then combined by dot product and then combined with the weights assigned by the attention layer. Following equation \ref{eq2} the product can be formulated as:

\begin{align}
    \textbf{\textit{h\textsubscript{t}}} = [\overrightarrow{h\textsuperscript{T}\textsubscript{i}}, \overleftarrow{h\textsuperscript{T}\textsubscript{i}}]\textsuperscript{T}
\end{align}

The \textbf{\textit{h}\textsubscript{i}} is then with the attention weights $\alpha$. The output vector is passed into the decoder where simple RNN aligns the sequence of the encoder output with \textbf{\textit{y\textsubscript{t-1}}}, eq \ref{eq3}.


\begin{figure}[!htb]
\centering
\centering
     \includegraphics[width=0.8\linewidth]{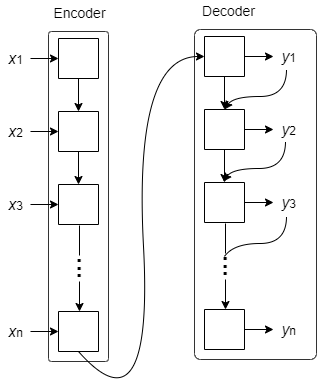}
     \caption{\label{Fig:image05}A high level representation of attention model with recurrent layers.}
\end{figure}

\section{Corpora}
English is considered to be a high resource language. There are many datasets available for English to work with. For our research, we used an English dictionary and tweets in the English language scrapped from Twitter. Roman Urdu is used for daily chat and tweet. There are a few datasets publicly available but they are not enough. So to increase our dataset size we targeted both Aryan Urdu and Roman Urdu tweets. The Aryan Urdu is later converted into Roman Urdu. 

\vspace{5pt}
\noindent\textbf{Dataset Preparation}:

Since we are collecting data from Twitter. The problem with data on social media is that it is highly informal with a lot of misspellings, characters lengthening, and spelling variations. Some words are commonly used in both English and Urdu languages. Thus, these challenges make code-mixing identification a difficult task. An example of these challenges can be found in Table~\ref{challenges-cm}.

\begin{figure}[!htb]
     \centering
     \includegraphics[width=0.8\linewidth]{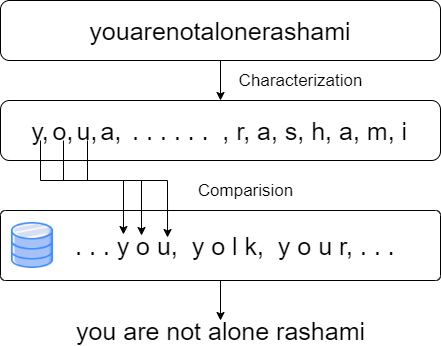}
     \caption{\label{Fig:image01}The image shows the process of converting a spaceless sentence into words chunks based on high matching sequence in both existing en and ru dictionaries.}
\end{figure}

To prepare data the very first step we took was to normalize words. Now there were sentences, in the scrapped data, which had no spaces between them. Using the pre-scraped English and Roman Urdu datasets we split the word into a series of characters and then compared the sequence with the words in the dictionary. The maximum sequence of characters which resembles a word in the dictionary was treated as a single word unit. Figure~\ref{Fig:image01} shows the process of dissecting a whole sentence into words. We performed tokenization on the sentences and then removed duplicates from the data.

\vspace{1.8pt}

\begin{table}
\centering
\begin{tabular}{ll}
\hline \textbf{CM Challenges} & \textbf{Examples}\\ \hline
Misspellings &  hat - hot (en)\\
Character lengthening & youuuu (en) \\ & kiaaa (ru) \\
Spelling variations &  thank you (thanku, \\ & tankyu) (en) \\ & (kia, kiya) (ru)\\
Common words & meh school(en, ru) jaa \\& raha hu\\
\hline
\end{tabular}
\caption{\label{challenges-cm} Bilingual Code Mixing challenges in languages with examples. The en tag represents English language while ru represents Roman Urdu word.}
\end{table}

\noindent\textbf{Dataset Annotation}:

The dataset is annotated into three classes i.e. English (en), Roman Urdu (ru), and rest (rs) which contains the information of punctuations or name entities. The data annotation process is done in a semi-supervised manner. When we are tackling with Roman Urdu and in this type of mixing usually every person has different formats for writing a single word. This is because there is no predefined standard for writing Urdu in the Roman version so everyone can have different combinations for characters for a single word as they deem fit.

Most of the words in this problem suffer from a character lengthening problem. To resolve this problem we can easily annotate the words written in Roman Urdu by maintaining a dictionary predefined dictionaries and with an addition to the scrapped words by reducing the characters lengthening to a single character. Hence, this technique will produce a basic form. Using this form we can simply find out other versions of the same word and annotate them. This technique also allows us to tackle the word duplication problems. For a single word, there can be many duplicates in spellings. We are not discarding the variations of spellings for a word because our goal is to pay attention to the words pronunciation representation of words in the vector space.

\begin{table}[!htb]
\centering
\begin{tabular}{lrr}
\hline \textbf{Class} & \textbf{EN Tokens} & \textbf{RU Tokens}\\ \hline
Train &  0.3458 & 0.0392\\
Dev &   0.0494 & 0.247\\
Test &  0.0988 & 0.0049\\
\hline
\end{tabular}
\caption{\label{table-datastats}The table shows the numbers of data tokens splitted in order to train and test the models.}
\end{table}

\vspace{5pt}
\noindent\textbf{Data Statistics}:

The dataset is randomly split into train and test parts as shown in Table~\ref{table-datastats}. We can see that the English dataset exceeds the Roman Urdu dataset, but many words are used the same in both languages as shown in the example~\ref{example_1}. 

\begin{quote}
    \label{example_1}\textbf{Example:} Shampoo, Jungle, Loot, University etc.
\end{quote}

For the English dataset at first, we used dictionary words which constitutes of 466k words\footnote{\url{https://github.com/dwyl/english-words}}. People usually use informal language in chats so we used a pre-trained chat dataset in English with about 1 Million sentences. For Roman, Urdu work has been done for semantical analysis of tweets. Different datasets are publicly available on Kaggle and Github repositories. We used those datasets as well as we used scrapped data from Twitter in both Roman Urdu and English.

For Roman Urdu, we also targeted tweets in Arayan Urdu. We then converted the Aryan Urdu into Roman Urdu by simply using machine transliteration API \footnote{\url{https://www.ijunoon.com/transliteration/urdu-to-roman}} from Arayan to Roman Urdu. Complete statistics about the dataset are shown in Table~\ref{data_stats}. 

\begin{table}
\centering
\begin{tabular}{lrl}
\hline \textbf{Dataset} & \textbf{Data size} & \textbf{Category}\\ \hline
 English Vocabulary& 0.374 & Words \\
 English Chat& 0.12& Words \\
 Roman Urdu& 0.049& Words \\
\hline
\end{tabular}
\caption{\label{data_stats}Dataset is the resource collected from different sources as well as the data which we scraped from Twitter. The data size is measured in millions of lines of words as per the category.}
\end{table}

After pre-processing, cleaning and annotating data the total dataset contains about 0.5 M of English tokens and 0.05 M of Roman Urdu tokens. 

\section{Experimental Setup}

For the validation of our research, we used two different systems. For data processing and collection we used a core i5 4210M CPU 2.40 GHz with 16 GB of RAM and an integrated Intel graphics card. To train our network we used a custom made AMD Ryzen\textsuperscript{TM 9} CPU, 16 GB of dedicated CUDA enabled NVIDIA GeForce RTX Super graphics card, 16 GB of RAM, and SSD NVMe storage. We used pycharm IDE for experimentation and the operating system used was Linux mint. 

\subsection{Models}

In this section, we will discuss some baseline models that were previously for multilingual language detection. We will then compare the accuracy of these models with the proposed model for this job.

\vspace{5pt}
\noindent\textbf{Hidden Markov Model (HMM)}

\begin{figure}[!htb]
     \centering
     \includegraphics[width=0.8\linewidth]{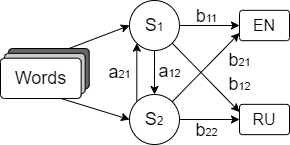}
     \caption{\label{Fig:image02}The architecture of Hidden Markov Model in which the connection with label \emph{a} shows the hidden states of the model while \emph{b} shows the output states.}
\end{figure}

Hidden Markov Model (HMM) is used in many sequence related problems. HMM is a statistical model which works on the probability of state switching between states for against provided sequence. Let say we have set of classes C = \{c\textsubscript{1}, c\textsubscript{2}, . . . , c\textsubscript{n}\}, which in our case are two i.e. en and ru. Then for \emph{d} dimensional sequence S = \{s\textsubscript{1}, s\textsubscript{2}, . . . , s\textsubscript{d}\}. The HMM model assigns the class to a word by calculating the forward and backward probabilities of classes along with the sequence provided as an input. The process is shown in the  Figure~\ref{Fig:image02}.

\vspace{5pt}
\noindent\textbf{Conditional Random Field (CRF)}

\begin{figure}[!htb]
     \centering
     \includegraphics[width=0.8\linewidth]{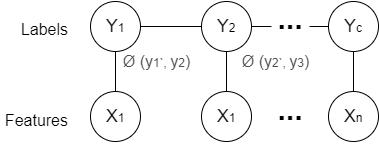}
     \caption{\label{Fig:image04}Architecture of Conditional Random Field.}
\end{figure}

Conditional Random Field (CRF) also a statistical model which predicts class \emph{C} for a given input of a sequence and based on the probability it assigns the class to the sequence \emph{S} \cite{lafferty2001conditional}. The Hidden Markov Model is doing the same then how CRF is different? But CRF is a discriminative model working on the principles of conditional probability distribution while HMM is based on joint probability distribution. Figure~\ref{Fig:image04} explains the working of CRF in context with bilingual word detection. 

\begin{table*}
\centering
\begin{tabular}{llllll}
\hline \textbf{Models} & \textbf{Accuracy} & \textbf{Precision} & \textbf{Recall} & \textbf{F1 Score}\\ \hline
 Hidden Markov Model& 0.85 & 0.73 & 0.81 & 0.72\\
 Conditional Random Field & 0.89 & 0.79 & 0.85  & 0.81\\
 Bi-directional LSTM & 0.92 & 0.82 & 0.71  & 0.79\\
 Attention Model& 0.93 & 0.87 & 0.75 & 0.84\\
\hline
\end{tabular}
\caption{\label{data_results}The table shows the evaluation metrics for different models we used in our paper in comparision to the attention model.}
\end{table*}

\vspace{5pt}
\noindent\textbf{Bidirectional LSTM}

A bidirectional LSTM \cite{zhou2016attention} we have two LSTM architecture. The input sequence is first passed into the LSTM in the default sequence and for the other LSTM model, the input is passed in reverse order. The results of which are combined. The weights are assigned with the attention mechanism. The results are passed through some activation function to get some input. The process is explained on a running example in Figure~\ref{Fig:image03} below.

\begin{figure}[!htb]
     \centering
     \includegraphics[width=1\linewidth]{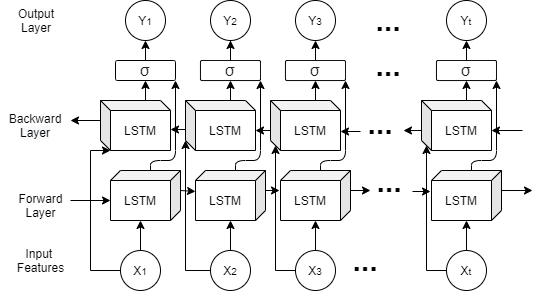}
     \caption{\label{Fig:image03}The architecture shows the workflow of a bidirectional LSTM. The first layer pass the sequence of features from 1 to n then the backward pass the sequence of features are passed from n to 1 then based on weight and activation function we get the output class.}
\end{figure}

\section{Results}

The annotated data used for experimentation can be used in many ways for multilingual analysis. We can observe the points to pay attention to which a speaker switches to the other language or what is the ratio for each language in code-switching. To evaluate language identification at the word level we use the following metrics. 

We start from the basic measuring metrics for the evaluation of any classification task. This is precision, accuracy and recall. These metrics give us an overall overview of the document. It can provide useful information about the user preference of language on the internet. Finally, we will use the F1 score to evaluate the accuracy of the model as compared to previous models. 

The comparison of the results between each model can be compared shown in Table~\ref{data_results}. The datasets from different sources were combined for training purposes. A significant amount of tests was done by comparing the results. For basic machine learning-based models it showed that Conditional Random Field (CRF) outperformed Hidden Markov Model (HMM). It can be seen that F1-Score is higher for CRF than HMM.

The bidirectional LSTM gives an accuracy of 0.92. Using the attention model the accuracy moves to about 0.93 but the main comparison can be compared from the F1 score, precision and recall. It is evident by the comparison of results that the attention model outperformed the results from a bidirectional LSTM model. Using the attention layer we observed that time consumed by it was comparatively low from traditional RNN Neural Network. For a simple RNN, we faced the problem of memory full issue.

\section{Conclusion}

In this paper, we observed that using attention improves the results and reduce the steps of a network. Attention networks can be proved useful in handling different tasks of low resource language processing. However, the dataset for roman Urdu was overwhelmed by the English data. In the future, we can compare results by combining different datasets to see what type of data gives good results. 

\section{Future Work}
Today we can see much work done in machine transliteration from one language to any other targeted language. As most of the internet communication is done in text code mixing is a common phenomena. We see room to contribute in the development of an architecture which does machine transliteration in terms of code mixed text. The basic idea is to take code mixed data and preserving the context of sentence translate into targeted language. This will be help full to extract useful semantics from code mixed text on internet. These semantics will be proved to be useful for the business community to improve there services based on customer reviews.
\cite{*}
\bibliography{anthology,acl2020}
\bibliographystyle{acl_natbib}

\end{document}